\documentclass[10pt]{article}
\usepackage[preprint]{tmlr}
\usepackage{booktabs}
\usepackage{array}
\usepackage{graphicx}
\usepackage{amsmath}
\usepackage{microtype}
\usepackage{tikz}
\usetikzlibrary{positioning,arrows.meta}
\usepackage{float}
\usepackage{hyperref}
\usepackage{url}
\hypersetup{hidelinks}
\title{SemPlan: Benchmarking Structured Semantic Planning for LLM-Based Queries over Enterprise Data\thanks{Generative AI tools assisted development and debugging of the experimental software pipeline. The author is responsible for the study design, validation, analysis, and scientific claims.}}
\author{\name Bruno Santos Teixeira \\
\addr Universidade Federal de Ouro Preto (UFOP), Brazil \\
\addr ORCID: \href{https://orcid.org/0009-0007-3860-7114}{0009-0007-3860-7114}}
\def\month{08}
\def\year{2026}
\def\openreview{\url{}}

\begin{document}
\maketitle
\begin{abstract}
Natural-language interfaces to enterprise data must translate underspecified requests into governed, executable behavior while controlling invalid queries, policy failures, cost, and nondeterminism. SemPlan Benchmark evaluates this architectural design space with a deterministic synthetic bilingual benchmark containing 1,800 cases in English and Brazilian Portuguese; 1,200 cases form the frozen scientific evaluation subset. Four architectures are compared under the same model configuration: direct SQL generation (A1), a bounded tool-agent baseline (A2), structured semantic-request generation followed by deterministic planning and execution (A3), and a clarification/stateful semantic-plan variant (A4). Across 4,800 primary records, answer correctness was low in absolute terms: 22.25\% for A1, 22.58\% for A2, 25.67\% for A3, and 24.25\% for A4. A3 had the highest observed correctness and significantly exceeded A1, A2, and A4 in the pre-specified paired correctness analysis, while A1 retained the highest policy-correct rate and the lowest unsafe-or-invalid rate. A4 had the lowest mean API cost and lowest false-refusal rate. On a preselected 150-case stability subset, answer-correct repeatability ranged from 92.00\% to 98.67\%. The results support a trade-off interpretation rather than a universal ranking: additional structural constraints changed failure modes and efficiency, but did not monotonically improve correctness or solve ambiguity and multi-turn state consistency.
\end{abstract}

\section{Introduction}
Natural-language database access has long promised to make structured data available to users who do not write query languages. The core difficulty, however, is not merely generating syntactically valid SQL. Enterprise analytics questions often refer to business metrics, implicit dimensions, temporal scopes, authorization boundaries, unsupported actions, and conversational context. A system can therefore be syntactically correct while still being semantically wrong, policy-inconsistent, or operationally invalid. Classical natural-language interfaces to databases identified ambiguity, domain modeling, and portability as persistent concerns \citep{androutsopoulos1995nlidb,affolter2019survey}; modern benchmarks such as Spider, SParC, CoSQL, and BIRD have renewed these questions under cross-domain, conversational, and database-grounded settings \citep{yu2018spider,yu2019sparc,yu2019cosql,li2023bird}.

Current LLM-based systems can place different amounts of responsibility on the model. At one extreme, the model generates SQL directly. Other designs ask the model to select tools, generate a typed semantic representation, or interact with deterministic normalization and execution layers. These choices are often discussed as engineering preferences, but they also define distinct empirical hypotheses about where errors arise and which constraints improve reliability.

SemPlan studies this design space while holding the model, benchmark cases, database snapshot, and evaluation rules fixed. The central question is: \emph{how do different forms and degrees of structural constraint in LLM-mediated enterprise-data querying trade off answer correctness, policy behavior, failure modes, cost, and repeatability?} The study is deliberately narrower than a claim of production readiness. It uses an independently generated synthetic enterprise domain and does not use customer or proprietary data.

The paper makes four contributions. First, it introduces a bilingual synthetic benchmark for controlled evaluation of LLM-mediated structured enterprise-data queries. Second, it compares four architecture patterns under a common experimental configuration rather than comparing different model families. Third, it reports paired evidence on answer correctness, policy behavior, typed failure outcomes, cost, ambiguity, multi-turn state consistency, and repeatability. Fourth, it provides a reproducibility package built around frozen prompts, manifests, raw-output hashes, score records, generated tables and figures, and no-key validation paths.

\section{Related Work}
Natural-language interfaces to databases predate current LLM systems and have historically relied on grammar, parsing, semantic representations, domain constraints, and schema-specific transformations \citep{androutsopoulos1995nlidb,affolter2019survey}. Modern text-to-SQL research shifted toward cross-domain evaluation. Spider established a challenging setting with databases and query structures that differ between training and test data \citep{yu2018spider}. SParC extended the task to contextual interactions, while CoSQL added conversational phenomena such as clarification and unanswerable requests \citep{yu2019sparc,yu2019cosql}. BIRD further emphasized large database contents, external knowledge, and efficiency \citep{li2023bird}.

A second line of work reduces the gap between language and executable SQL through intermediate structure or constrained decoding. IRNet uses schema linking and a SemQL intermediate representation before deterministic SQL inference \citep{guo2019irnet}. RAT-SQL emphasizes relation-aware schema encoding and linking \citep{wang2020ratsql}. Execution-guided decoding filters candidates using execution feedback, while PICARD incrementally rejects inadmissible tokens during generation \citep{wang2018executionguided,scholak2021picard}. These methods motivate the broader principle that formal constraints can reduce some classes of invalid output, but they do not imply that additional constraints always improve end-to-end task correctness.

Tool-using language-model systems represent another architecture family. ReAct interleaves reasoning and actions, while Toolformer studies how models decide when and how to invoke external APIs \citep{yao2023react,schick2023toolformer}. In current API systems, structured outputs and function-calling mechanisms can constrain model responses or tool arguments to schemas \citep{openai2026structuredoutputs,openai2026functioncalling}. SemPlan is complementary to these lines of work: rather than proposing a single decoding technique, it evaluates four end-to-end architectural allocations of responsibility between an LLM and deterministic software components.

The benchmark design also follows the broader reproducibility principle that datasets should communicate motivation, composition, generation, intended use, and limitations \citep{gebru2021datasheets}. SemPlan therefore separates synthetic data generation, benchmark generation, human review, gold execution, model execution, and derived statistical artifacts.

\section{Research Questions and Pre-Specified Hypotheses}
The study was frozen around five research questions. \textbf{RQ1} asks how A1--A4 differ in final answer correctness on bilingual enterprise-analytics questions. \textbf{RQ2} asks how they differ in invalid, execution-failure, and policy-related outcomes. \textbf{RQ3} asks what cost and latency trade-offs arise. \textbf{RQ4} asks whether formal clarification improves behavior on materially ambiguous questions. \textbf{RQ5} asks whether structured state improves consistency in PATCH/REPLACE follow-ups.

Before the hidden evaluation, directional hypotheses expected semantic-plan approaches to improve semantic and answer correctness over direct SQL; to reduce invalid or policy-violating execution relative to less structured alternatives; to keep A3 cost at or below the tool-agent approach; and to improve ambiguity and multi-turn consistency through A4 clarification and structured state. The analysis below preserves mixed and negative findings rather than redefining those hypotheses after observing results.

\section{SemPlan Benchmark}
\subsection{Synthetic enterprise domain}
The benchmark uses the synthetic \emph{Northstar Commerce} analytics domain. Its relational data cover customers, products, orders, payments, expenses, budgets, suppliers, contracts, and a calendar dimension. Monetary values, dates, status fields, and derived business metrics are generated from deterministic seeds and documented invariants. The reference execution environment uses governed PostgreSQL views and a read-only execution policy.

\subsection{Case generation, review, and splits}
Benchmark construction separates textual cases from the underlying tabular dataset. Parameterized semantic templates define intents, metrics, dimensions, filters, temporal contexts, grouping, sorting, limits, ambiguity conditions, and policy expectations. Surface forms are generated for English and Brazilian Portuguese, then validated for canonical identifiers and language quality. Gold semantic plans execute deterministically against the reference database to produce normalized gold answers. Cases are deduplicated by structural and lexical lineage before the test partitions are frozen.

The release-scale benchmark contains 1,800 cases: 900 en-US and 900 pt-BR. Its splits are development (300), validation (300), public test (500), hidden test (300), multi-turn (200), and adversarial (200). The frozen scientific evaluation subset contains 1,200 cases: public test, hidden test, multi-turn, and adversarial. Development and validation cases are excluded from the primary scientific comparison. The release manifest records 3,600 approved case/gold review records from a single human reviewer. This strong author-in-the-loop role is treated as a validity limitation rather than hidden.

\begin{table}[H]
\centering
\caption{Composition of the frozen SemPlan Benchmark version 1.0.0-rc.2.}
\label{tab:composition}
\small
\renewcommand{\arraystretch}{1.06}
\begin{tabular}{@{}llr@{}}
\toprule
\textbf{Dimension} & \textbf{Value} & \textbf{Cases} \\
\midrule
\textbf{Split} & Adversarial & 200 \\
 & Development & 300 \\
 & Multi-turn & 200 \\
 & Hidden test & 300 \\
 & Public test & 500 \\
 & Validation & 300 \\
\addlinespace[0.35em]
\textbf{Language} & English (en-US) & 900 \\
 & Portuguese (pt-BR) & 900 \\
\addlinespace[0.35em]
\textbf{Difficulty} & Easy & 236 \\
 & Medium & 840 \\
 & Hard & 724 \\
\bottomrule
\end{tabular}

\end{table}

The benchmark includes lookup, grouped aggregation, ranking, comparison, variance, trend, share/ratio, filtering, contract-status, ambiguity, out-of-scope, multi-turn, and adversarial cases. Difficulty labels derive from documented structural factors rather than model performance.

\subsection{Illustrative bilingual case}
A typical single-turn case asks for a governed metric under explicit business filters. For example, an English utterance may be \emph{``What was net revenue for consumer customers in the North region through the online channel in January 2022?''}; its Brazilian-Portuguese counterpart can express the same semantics as \emph{``Qual foi a receita l\'iquida dos clientes do segmento consumidor na regi\~ao Norte pelo canal online em janeiro de 2022?''}. The canonical semantic request identifies \texttt{net\_revenue}, the region/channel/segment filters, and the time window. This example illustrates the benchmark contract; primary evaluation uses the frozen case artifacts rather than manually rewritten examples in the manuscript.

\section{Evaluated Approaches}
\begin{figure}[H]
\centering
\resizebox{0.98\linewidth}{!}{\begin{tikzpicture}[
    >=Stealth,
    box/.style={
        draw,
        rounded corners=1.5pt,
        align=center,
        minimum height=0.86cm,
        text width=2.25cm,
        inner sep=3.5pt,
        font=\small
    },
    widebox/.style={
        draw,
        rounded corners=1.5pt,
        align=center,
        minimum height=0.86cm,
        text width=2.75cm,
        inner sep=3.5pt,
        font=\small
    },
    dbbox/.style={
        draw,
        rounded corners=1.5pt,
        align=center,
        minimum height=0.86cm,
        text width=1.05cm,
        inner sep=3.5pt,
        font=\small
    },
    arr/.style={->, line width=0.75pt},
    rowlabel/.style={font=\bfseries\large, anchor=east}
]

\node[rowlabel] (a1l) at (0,0) {A1};
\node[box]     (a1q) at (1.65,0) {User query};
\node[box]     (a1m) at (4.45,0) {LLM};
\node[box]     (a1o) at (7.25,0) {SQL};
\node[widebox] (a1x) at (10.35,0) {SQL guard};
\node[dbbox]   (a1d) at (12.95,0) {DB};
\draw[arr] (a1q.east) -- (a1m.west);
\draw[arr] (a1m.east) -- (a1o.west);
\draw[arr] (a1o.east) -- (a1x.west);
\draw[arr] (a1x.east) -- (a1d.west);

\node[rowlabel] (a2l) at (0,-1.65) {A2};
\node[box]     (a2q) at (1.65,-1.65) {User query};
\node[box]     (a2m) at (4.45,-1.65) {LLM agent};
\node[box]     (a2o) at (7.25,-1.65) {Typed tools};
\node[widebox] (a2x) at (10.35,-1.65) {Deterministic\\executor};
\node[dbbox]   (a2d) at (12.95,-1.65) {DB};
\draw[arr] (a2q.east) -- (a2m.west);
\draw[arr] (a2m.east) -- (a2o.west);
\draw[arr] (a2o.east) -- (a2x.west);
\draw[arr] (a2x.east) -- (a2d.west);

\node[rowlabel] (a3l) at (0,-3.30) {A3};
\node[box]     (a3q) at (1.65,-3.30) {User query};
\node[box]     (a3m) at (4.45,-3.30) {LLM};
\node[widebox] (a3o) at (7.25,-3.30) {SemanticRequest};
\node[widebox] (a3x) at (10.35,-3.30) {Normalize +\\execute};
\node[dbbox]   (a3d) at (12.95,-3.30) {DB};
\draw[arr] (a3q.east) -- (a3m.west);
\draw[arr] (a3m.east) -- (a3o.west);
\draw[arr] (a3o.east) -- (a3x.west);
\draw[arr] (a3x.east) -- (a3d.west);

\node[rowlabel] (a4l) at (0,-4.95) {A4};
\node[box]     (a4q) at (1.65,-4.95) {Query +\\structured state};
\node[box]     (a4m) at (4.45,-4.95) {LLM};
\node[widebox] (a4o) at (7.25,-4.95) {Request / clarify\\/ patch};
\node[widebox] (a4x) at (10.35,-4.95) {Normalize + state\\+ execute};
\node[dbbox]   (a4d) at (12.95,-4.95) {DB};
\draw[arr] (a4q.east) -- (a4m.west);
\draw[arr] (a4m.east) -- (a4o.west);
\draw[arr] (a4o.east) -- (a4x.west);
\draw[arr] (a4x.east) -- (a4d.west);

\end{tikzpicture}}
\caption{Where structure enters the SemPlan comparison. Only A1 permits model-authored SQL. A2 constrains model actions to typed tools. A3 and A4 emit typed semantic requests that are normalized and executed by deterministic software; A4 additionally exposes structured state and clarification behavior.}
\label{fig:architecture}
\end{figure}
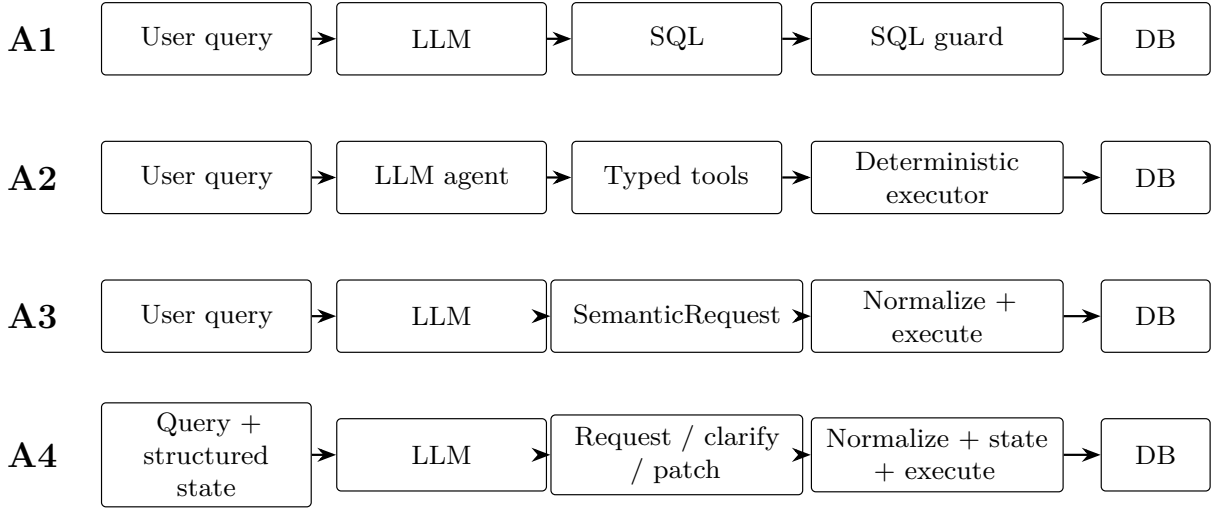

All four approaches use the same configured model, \texttt{gpt-5.6-luna}, through a provider-neutral OpenAI adapter with reasoning effort \texttt{low} and a 1,200-token output ceiling. Prompts are frozen independently per approach. Figure~\ref{fig:architecture} summarizes the allocation of model and deterministic responsibilities.

\textbf{A1: Direct SQL.} The model receives the utterance, locale, reference date, a compact governed view schema, and relevant metric definitions. It returns strict JSON containing SQL or a typed refusal. Generated SQL is not executed directly: it must pass parsing, AST allowlists, complexity limits, read-only authorization, statement timeout, and row limits. A1 never receives gold SQL, plans, or answers.

\textbf{A2: Tool Agent.} The model receives a bounded set of analytics tools for operations such as aggregation, ranking, period comparison, budget comparison, contract status, and field description. Tool arguments follow strict schemas and canonical catalog identifiers. The agent may compose only a bounded number of tool calls and cannot execute arbitrary SQL or invent tools. The deterministic executor implements the underlying analytics operations.

\textbf{A3: Semantic Request.} The model does not generate SQL. Instead, it emits a strict semantic-request envelope containing operation, metrics, dimensions, filters, time grain, sort, limit, comparison fields, clarification fields, and confidence. Deterministic normalization validates identifiers and types, resolves documented semantic rules, and compiles the request into an executable semantic plan. Governed software then produces parameterized SQL or an equivalent deterministic operator call. Thus SQL exists in A3, but it is generated by the application, not by the LLM.

\textbf{A4: Clarification and Structured State.} A4 extends the A3 interface with prior structured state, typed clarification outcomes, and explicit PATCH/REPLACE semantics for multi-turn requests. Only fields explicitly changed by a PATCH are updated subject to compatibility rules; out-of-scope turns preserve state unless reset. Execution remains deterministic after normalization.

\section{Experimental Setup}
The frozen experiment uses benchmark version 1.0.0-rc.2 and the same case set, database snapshot, model configuration, and approach-specific frozen prompts for all primary comparisons. Each of the 1,200 scientific cases is executed once under each of A1--A4, producing 4,800 primary records. A deterministic stratified subset of 150 cases is then executed two additional times per approach, producing 1,200 stability records. These repeats are analyzed only for within-approach repeatability and do not inflate the primary sample size. The complete scientific run therefore contains 6,000 planned request identities.

Primary binary endpoints are compared on case-aligned records. McNemar tests and paired risk differences with 95\% confidence intervals are used for binary outcomes, and Holm-Bonferroni adjustment is applied to the pre-specified primary comparison family. Skewed continuous outcomes such as API cost use paired bootstrap contrasts. Provider, transport, infrastructure, and model-quality failures are distinguished before statistical interpretation. Model-quality failures remain part of the scientific outcome when the system produced sufficient evidence to classify them.

Latency was pre-specified but is excluded from primary conclusions. The post-run audit found that captured typed failures record zero latency, and those failures are unevenly distributed across approaches. Consequently, the resulting medians are not comparable as a clean end-to-end latency measure. Token usage, provider cost, and failure counts remain reportable because they are independently recorded in provider usage and budget/result ledgers.

\section{Results}
\subsection{Primary correctness and policy outcomes}
The run completed all 6,000 planned request identities with zero missing identities and zero duplicate purchases. In the final state there were no provider failures and no uncaptured infrastructure failures. Table~\ref{tab:primary} reports the primary endpoints for the 1,200 cases per approach.

\begin{table}[t]
\centering
\small
\caption{Primary metrics over 1,200 scientific cases per approach. Absolute correctness is reported before pairwise significance.}
\label{tab:primary}
\begin{tabular}{lrrrrr}
\toprule
Approach & Answer correct & Policy correct & False refusal & Unsafe/invalid & Mean cost \\
\midrule
A1 & 22.25\% & 43.67\% & 21.42\% & 31.00\% & \$0.000918 \\
A2 & 22.58\% & 35.42\% & 1.83\% & 62.25\% & \$0.000927 \\
A3 & 25.67\% & 37.33\% & 1.58\% & 60.58\% & \$0.000512 \\
A4 & 24.25\% & 35.25\% & 0.17\% & 64.08\% & \$0.000469 \\
\bottomrule
\end{tabular}

\end{table}

\begin{figure}[t]
\centering
\includegraphics[width=0.82\linewidth]{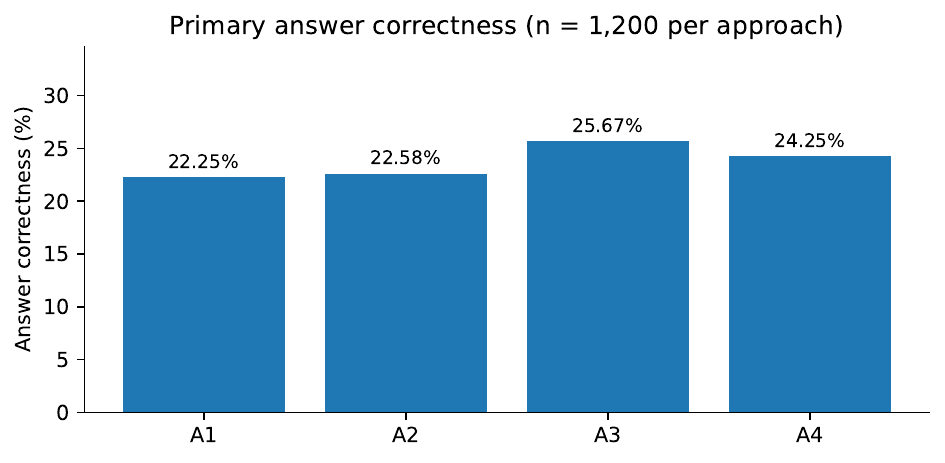}
\caption{Primary answer correctness from the frozen primary-results table. The figure shows absolute rates; paired uncertainty is reported in Table~\ref{tab:contrasts}.}
\label{fig:correctness}
\end{figure}

Absolute answer correctness is low across all approaches. A3 achieved the highest observed rate at 25.67\%, followed by A4 at 24.25\%, A2 at 22.58\%, and A1 at 22.25\% (Figure~\ref{fig:correctness}). The paired answer-correct analysis estimates A3 over A1 at +3.42 percentage points (95\% CI [1.85, 4.98], Holm-adjusted $p=0.000110$), and A3 over A2 at +3.08 points (95\% CI [1.68, 4.49], Holm-adjusted $p=0.000110$). A4 is 1.42 points below A3 (A4--A3 95\% CI [-2.51, -0.32], Holm-adjusted $p=0.041250$). Under the frozen analysis, A3 therefore has higher primary answer correctness than each of A1, A2, and A4, but the absolute rates remain modest.

Policy behavior yields a different ordering. A1 has the highest policy-correct rate (43.67\%) and the lowest unsafe-or-invalid rate (31.00\%). A4 has the lowest false-refusal rate (0.17\%) but the highest unsafe-or-invalid rate (64.08\%). A3 has 37.33\% policy correctness and 60.58\% unsafe-or-invalid. The paired policy analysis shows that A3 and A4 are both below A1 on policy correctness. The correctness advantage of A3 therefore does not establish a uniformly better or safer architecture.

\begin{table}[t]
\centering
\small
\caption{Selected paired binary contrasts. Risk difference is left minus right.}
\label{tab:contrasts}
\begin{tabular}{lrrrrr}
\toprule
Metric & Contrast & n & Risk diff. & 95\% CI & Holm p \\
\midrule
answer\_correct & A3 vs A1 & 1200 & +0.034167 & [+0.018534, +0.049799] & 0.000110 \\
answer\_correct & A4 vs A1 & 1200 & +0.020000 & [+0.004720, +0.035280] & 0.041250 \\
answer\_correct & A3 vs A2 & 1200 & +0.030833 & [+0.016796, +0.044870] & 0.000110 \\
answer\_correct & A4 vs A2 & 1200 & +0.016667 & [+0.003231, +0.030102] & 0.041250 \\
answer\_correct & A4 vs A3 & 1200 & -0.014167 & [-0.025094, -0.003239] & 0.041250 \\
policy\_correct & A3 vs A1 & 1200 & -0.063333 & [-0.090032, -0.036635] & 0.000020 \\
policy\_correct & A4 vs A1 & 1200 & -0.084167 & [-0.110223, -0.058110] & 0.000000 \\
policy\_correct & A3 vs A2 & 1200 & +0.019167 & [+0.001685, +0.036649] & 0.079502 \\
policy\_correct & A4 vs A2 & 1200 & -0.001667 & [-0.018000, +0.014666] & 0.920411 \\
policy\_correct & A4 vs A3 & 1200 & -0.020833 & [-0.034929, -0.006737] & 0.015684 \\
\bottomrule
\end{tabular}

\end{table}

\subsection{Exploratory language subgroup}
Language-stratified correctness is consistently higher on en-US than pt-BR for all four approaches (Table~\ref{tab:language}). A3 records 31.00\% in en-US and 20.33\% in pt-BR; the corresponding A4 rates are 29.33\% and 19.17\%. Because subgroup comparisons were not the sole primary endpoint and template families can be correlated, these results are descriptive/exploratory rather than evidence that the architecture causes a language-specific effect. They nevertheless identify multilingual robustness as an important follow-up question.

\begin{table}[t]
\centering
\caption{Exploratory primary answer correctness by language. Each language subgroup has $n=600$ cases per approach.}
\label{tab:language}
\begin{tabular}{lrr}
\toprule
Approach & en-US & pt-BR \\
\midrule
A1 & 24.67\% & 19.83\% \\
A2 & 28.50\% & 16.67\% \\
A3 & 31.00\% & 20.33\% \\
A4 & 29.33\% & 19.17\% \\
\bottomrule
\end{tabular}

\end{table}

\section{Failure Analysis}
Typed model-quality failures are central to the interpretation rather than discarded as missing data. Across all 6,000 scientific records, the recorded error codes are \texttt{EXECUTION\_FAILED} (1,762), \texttt{OUTPUT\_SCHEMA\_INVALID} (1,051), \texttt{POLICY\_VIOLATION} (368), \texttt{CATALOG\_UNKNOWN\_ID} (78), and \texttt{CFG\_INVALID} (24). In the 4,800 primary records alone, the corresponding counts are 1,402, 845, 287, 65, and 16.

\begin{table}[t]
\centering
\small
\caption{Typed model-quality failure categories. These are scientific outcomes rather than missing provider evidence.}
\label{tab:errorcodes}
\begin{tabular}{lrr}
\toprule
Error code & All records & Primary records \\
\midrule
EXECUTION\_FAILED & 1762 & 1402 \\
OUTPUT\_SCHEMA\_INVALID & 1051 & 845 \\
POLICY\_VIOLATION & 368 & 287 \\
CATALOG\_UNKNOWN\_ID & 78 & 65 \\
CFG\_INVALID & 24 & 16 \\
\bottomrule
\end{tabular}

\end{table}

\texttt{EXECUTION\_FAILED} indicates that a model-produced or normalized request reached governed execution but failed a typed database/execution rule. \texttt{OUTPUT\_SCHEMA\_INVALID} indicates failure of the strict output contract. \texttt{POLICY\_VIOLATION} records guard-layer rejection; \texttt{CATALOG\_UNKNOWN\_ID} records unknown semantic identifiers; and \texttt{CFG\_INVALID} records a captured configuration or contract invalidity.

Approach-level terminal outcomes show that additional structure did not simply remove errors. A1 produced 372 primary \texttt{ERROR} outcomes, compared with 747 for A2, 727 for A3, and 769 for A4. At the same time, A1 produced 257 false refusals, compared with 22, 19, and 2 for A2--A4. This pattern supports a failure-shift interpretation: direct SQL plus strict guards is conservative in one way, while typed semantic/tool interfaces expose different contract and execution failure surfaces. The aggregate artifacts do not establish a unique causal mechanism for each error, so the discussion avoids attributing individual error-code differences to specific modules without case-level analysis.

\begin{table}[t]
\centering
\small
\caption{Primary terminal-outcome counts by approach.}
\label{tab:failures}
\begin{tabular}{lrrrrr}
\toprule
Approach & Answered & Clarify & Out-of-scope & Error & False refusal \\
\midrule
A1 & 292 & 0 & 536 & 372 & 257 \\
A2 & 210 & 29 & 214 & 747 & 22 \\
A3 & 194 & 33 & 246 & 727 & 19 \\
A4 & 194 & 9 & 228 & 769 & 2 \\
\bottomrule
\end{tabular}

\end{table}

\section{Stability, Ambiguity, and Multi-Turn Behavior}
The stability substudy uses the primary execution plus two additional independent executions for each of 150 preselected cases per approach. Repeatability asks whether answer correctness is stable across those three executions; it is not a claim of full determinism. A3 has the highest observed repeatability (98.67\%), A1 and A4 are both 98.00\%, and A2 is 92.00\%.

\begin{table}[t]
\centering
\caption{Answer-correct repeatability on the 150-case stability subset.}
\label{tab:stability}
\begin{tabular}{lrrr}
\toprule
Approach & Repeatability & Cases & Additional records \\
\midrule
A1 & 98.00\% & 150 & 300 \\
A2 & 92.00\% & 150 & 300 \\
A3 & 98.67\% & 150 & 300 \\
A4 & 98.00\% & 150 & 300 \\
\bottomrule
\end{tabular}

\end{table}

Conversational robustness is substantially weaker. Clarification-decision correctness is 0.00\% for A1, 26.19\% for A2, 27.38\% for A3, and 8.33\% for A4 over 84 ambiguity cases per approach. Multi-turn sequence-state correctness is 20.00\%, 17.00\%, 14.00\%, and 14.50\% for A1--A4 over 200 multi-turn cases. The clarification-enabled A4 design therefore does not validate the pre-specified expectation that structured clarification would dominate the alternatives, and structured state does not improve sequence-state correctness in this configuration.

\begin{table}[t]
\centering
\small
\caption{Ambiguity and multi-turn endpoints. Weak absolute rates are treated as primary scientific limitations.}
\label{tab:ambiguity}
\begin{tabular}{lrrrr}
\toprule
Approach & Clarification correct & Ambiguity n & Sequence-state correct & Multi-turn n \\
\midrule
A1 & 0.00\% & 84 & 20.00\% & 200 \\
A2 & 26.19\% & 84 & 17.00\% & 200 \\
A3 & 27.38\% & 84 & 14.00\% & 200 \\
A4 & 8.33\% & 84 & 14.50\% & 200 \\
\bottomrule
\end{tabular}

\end{table}

\section{Cost and Operational Results}
The complete F7 scientific run recorded USD 4.028973 in API charges. Mean cost per primary record is USD 0.000918 for A1, 0.000927 for A2, 0.000512 for A3, and 0.000469 for A4. A4 is the least expensive observed approach and A3 is also substantially cheaper than A1/A2 under this frozen model and price table. The paired mean cost difference is -USD 0.000415 for A3 versus A2 and -USD 0.000459 for A4 versus A2. These values are provider-API observations, not total cost of ownership; they exclude infrastructure, engineering time, and any future pricing changes.

\begin{figure}[t]
\centering
\includegraphics[width=0.82\linewidth]{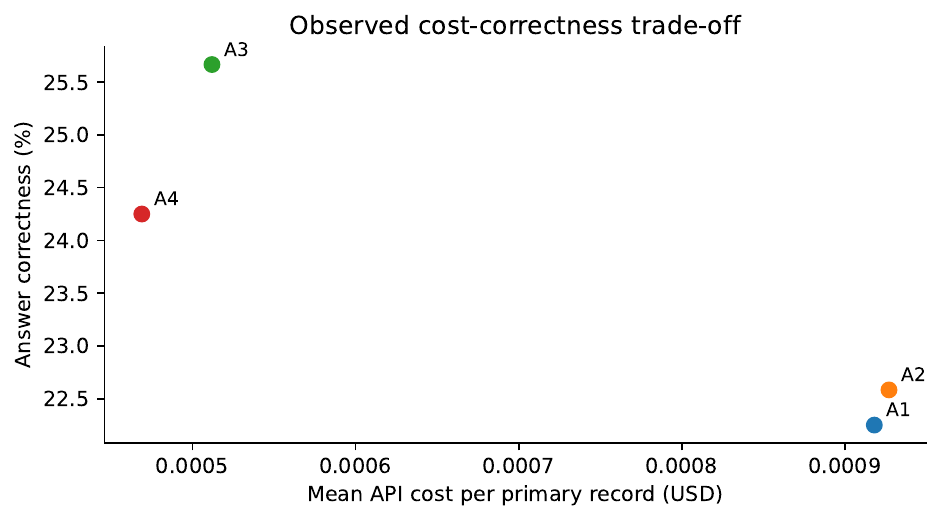}
\caption{Observed cost-correctness trade-off from frozen primary tables. A3 has the highest correctness, while A4 has the lowest mean API cost.}
\label{fig:frontier}
\end{figure}

Total recorded provider usage is 8,120,408 input tokens, 1,179,317 cached input tokens, 1,540,985 output tokens, and 639,950 reasoning tokens. These counts are useful for reproducibility and budget analysis but should not be generalized to other models or future pricing.

\section{Discussion}
The main result is not a monotonic ``more structure is better'' relationship. A3 occupies the strongest observed correctness/repeatability position: it has the highest answer correctness, the highest stability repeatability, and a mean cost well below A1/A2. Yet A3 does not have the strongest policy behavior and its unsafe-or-invalid rate remains high. A4 adds structured state and clarification, achieves the lowest mean API cost and almost eliminates false refusals, but it is significantly below A3 on answer correctness and has the highest unsafe-or-invalid rate. A1 is comparatively weak on answer correctness and false refusal, yet it has the best policy-correct rate and the lowest unsafe-or-invalid rate. A2 yields only a small raw correctness increase over A1, has the weakest repeatability, and a large error surface.

A plausible architectural interpretation is that direct SQL gives the model a simpler but more permissive formal target, after which strict SQL guards reject unsafe behavior. A3 removes model-authored SQL and asks the model to solve a narrower semantic parsing problem; this can reduce some implementation freedom while retaining a compact interface. A4 introduces additional state and clarification semantics, which may create useful governance capabilities but also more opportunities for an otherwise useful request to become incompatible with a strict contract. These mechanisms are hypotheses for follow-up analysis, not causal conclusions established by aggregate rates.

The bilingual subgroup result is also important. Every approach is less accurate on pt-BR than en-US in the frozen primary set, even though the corpus is balanced by language and passed the release language-quality gate. This gap may reflect differences in surface-form difficulty, lexical mappings, model behavior, or interactions between those factors. It motivates independent multilingual review and additional model families rather than post-hoc retuning of the frozen benchmark.

Finally, the low absolute correctness rates are themselves a result. Even the best architecture is correct on roughly one quarter of primary cases under the benchmark's combined clean, adversarial, multi-turn, and policy-sensitive distribution. SemPlan should therefore be read as evidence about comparative architectural trade-offs under a difficult controlled benchmark, not as certification that any tested design is deployment-ready.

\section{Threats to Validity}
\textbf{Internal validity.} Prompt design, strict output contracts, implementation details, provider nondeterminism, retry/recovery behavior, and the F7 execution hotfix lineage can affect comparisons. The execution hotfix changed failure capture rather than benchmark semantics: database exceptions that previously could crash the runner were converted into typed scientific outcomes, preserving the underlying model/approach failure. The shared model, database snapshot, cases, and execution limits reduce but do not eliminate implementation asymmetry.

\textbf{Construct validity.} Answer correctness is defined by equivalence to a normalized canonical result; it is not a complete measure of user usefulness, explanation quality, or business trust. Policy correctness and unsafe-or-invalid rates cover the benchmark's specified policy outcomes, not every enterprise governance concern. Repeatability measures stability of a binary correctness outcome across three executions on a subset and is not equivalent to deterministic generation. Latency is explicitly excluded from primary claims because its instrumentation assigns zero to captured typed failures.

\textbf{External validity.} The study uses one synthetic enterprise domain, one model/provider configuration, two languages, PostgreSQL, and a controlled semantic catalog. Generalization to other schemas, data distributions, organizations, model families, providers, languages, or production traffic is unknown. No real customer data are used.

\textbf{Statistical conclusion validity.} The case-aligned paired design improves sensitivity for architecture comparisons, and the primary binary family uses multiplicity correction. Nevertheless, finite subgroup sizes, correlated template families, and multiple exploratory breakdowns constrain interpretation. Language, class, and difficulty analyses should not be promoted to new primary hypotheses after observing the data.

\textbf{Researcher validity.} A single human researcher designed the project and served as the primary gold reviewer. Automated gold execution, deterministic manifests, review queues, frozen hashes, and public reproducibility artifacts mitigate but do not remove author-in-the-loop bias. Independent replication and secondary annotation would strengthen future versions.

\section{Limitations}
SemPlan does not fine-tune models, compare multiple model families, evaluate production user interfaces, or certify deployment safety. The synthetic domain is designed to exercise enterprise-analytics semantics but cannot reproduce the full organizational context of real data systems. The study also does not claim that the four architectures exhaust the design space. In particular, hybrid systems could combine A3-style semantic requests with stronger policy enforcement or alternative state models. Such variants should be evaluated as new, pre-specified experiments rather than retrofitted into the completed F7 run.

Raw provider responses are hash-recorded locally. Public release of those raw responses should occur only after a final policy/licensing review; derived score records, frozen tables, figures, benchmark data, prompts, schemas, and deterministic generation/reproduction paths can be released independently. The current paper also avoids using latency as a comparative performance claim because of the instrumentation issue described above.

\section{Reproducibility and Ethics}
All benchmark data are synthetic and independently created. Code is licensed under Apache-2.0; benchmark/dataset documentation is licensed under CC BY 4.0 unless a file states otherwise. The release candidate includes schemas, catalog files, frozen prompts, deterministic generators, benchmark manifests, result tables, figures, audit documents, and checksums. A no-key validation path can regenerate deterministic data and derived paper artifacts without additional paid provider calls. The frozen scientific-result manifest binds raw-response, result-record, score-record, figure, table, and cost-ledger hashes.

The software-development use of generative AI is disclosed on the first page. No AI system is an author. Scientific design, validation, interpretation, claims, and publication decisions remain the responsibility of the human author.

\section{Conclusion}
SemPlan Benchmark compares four ways to divide responsibility between an LLM and deterministic software when answering enterprise-data questions. Under the frozen bilingual synthetic benchmark, A3 achieves the highest observed primary answer correctness and repeatability, A4 has the lowest mean API cost and false-refusal rate, and A1 retains the strongest policy-correct behavior and the lowest unsafe-or-invalid rate. The paired analysis supports a real but modest correctness advantage for A3, while the ambiguity, multi-turn, and policy results show that additional structure does not yield uniformly better behavior. These findings argue for evaluating architectural trade-offs along multiple dimensions rather than ranking systems by a single accuracy number. Future work should replicate the benchmark with additional model families, independent reviewers, other data domains and languages, and pre-specified variants that target the observed ambiguity, state, and contract-failure modes.

\section*{Artifact Availability}
This preprint is archived on Zenodo at DOI \href{https://doi.org/10.5281/zenodo.21904872}{\texttt{10.5281/zenodo.21904872}}. The scientific results are frozen under manifest SHA-256 \texttt{5f06bc8b...6aefb816}; full immutable identifiers are listed in Appendix~\ref{app:ids}. A public reproducibility package is being deposited separately, and dataset/software identifiers will be linked in future public versions.

\appendix
\section{Frozen Experiment Identifiers}\label{app:ids}
\begin{table}[h]
\centering
\footnotesize
\caption{Key immutable identifiers for the reported experiment.}
\begin{tabular}{lp{0.72\linewidth}}
\toprule
Artifact & Identifier \\
\midrule
Benchmark version & \texttt{1.0.0-rc.2} \\
Benchmark SHA-256 & \texttt{60fe39c51a71637afd2a88f9fd44c3eee620a6e5467591d3c36a0a39a16792ed} \\
Scientific results SHA-256 & \texttt{5f06bc8bad81a8c0c151d4890a34432bb0166e73331371fb5c7fb8536aefb816} \\
Model & \texttt{gpt-5.6-luna}, reasoning \texttt{low}, max output 1,200 \\
Primary design & 1,200 cases $\times$ 4 approaches $=$ 4,800 records \\
Stability design & 150 cases $\times$ 4 approaches $\times$ 2 additional runs $=$ 1,200 records \\
\bottomrule
\end{tabular}
\end{table}

\section{Frozen Prompt Hashes}
\footnotesize
\begin{tabular}{lp{0.78\linewidth}}
\toprule
Approach & Prompt SHA-256 \\
\midrule
A1 & \texttt{257c15c136c0a32108e41731a5d8b50c62c96b16e1ccdc2b973df92c6fad27d7} \\
A2 & \texttt{a50283334d378454192bd40d17a0747a225d3971a6203d8c9cc4443017293a05} \\
A3 & \texttt{7a5cb95ccd3efb42feb4bedf2177250cdb2ee866758e8fbf93e90c85abb8fc95} \\
A4 & \texttt{582d25398d5311e22b64d6472edb4e092dc85b095cc214acf86a1c37431cab51} \\
\bottomrule
\end{tabular}

\bibliographystyle{plainnat}
\bibliography{references}

@article{androutsopoulos1995nlidb,
  title = {Natural Language Interfaces to Databases - An Introduction},
  author = {Ion Androutsopoulos and Graeme D. Ritchie and Peter Thanisch},
  year = {1995},
  url = {https://arxiv.org/abs/cmp-lg/9503016},
  note = {arXiv:cmp-lg/9503016},
  journal = {Natural Language Engineering},
}

@article{affolter2019survey,
  title = {A Comparative Survey of Recent Natural Language Interfaces for Databases},
  author = {Katrin Affolter and Kurt Stockinger and Abraham Bernstein},
  year = {2019},
  url = {https://link.springer.com/article/10.1007/s00778-019-00567-8},
  note = {10.1007/s00778-019-00567-8},
  journal = {The VLDB Journal},
  doi = {10.1007/s00778-019-00567-8},
}

@inproceedings{yu2018spider,
  title = {Spider: A Large-Scale Human-Labeled Dataset for Complex and Cross-Domain Semantic Parsing and Text-to-SQL Task},
  author = {Tao Yu and Rui Zhang and Kai Yang and Michihiro Yasunaga and Dongxu Wang and Zifan Li and James Ma and Irene Li and Qingning Yao and Shanelle Roman and Zilin Zhang and Dragomir Radev},
  year = {2018},
  url = {https://aclanthology.org/D18-1425/},
  note = {10.18653/v1/D18-1425},
  booktitle = {EMNLP},
  doi = {10.18653/v1/D18-1425},
}

@inproceedings{yu2019sparc,
  title = {SParC: Cross-Domain Semantic Parsing in Context},
  author = {Tao Yu and Rui Zhang and Michihiro Yasunaga and Yi Chern Tan and Xi Victoria Lin and Suyi Li and Heyang Er and Irene Li and Bo Pang and Tao Chen and Emily Ji and Shreya Dixit and David Proctor and Sungrok Shim and Jonathan Kraft and Vincent Zhang and Caiming Xiong and Richard Socher and Dragomir Radev},
  year = {2019},
  url = {https://aclanthology.org/P19-1443/},
  note = {10.18653/v1/P19-1443},
  booktitle = {ACL},
  doi = {10.18653/v1/P19-1443},
}

@inproceedings{yu2019cosql,
  title = {CoSQL: A Conversational Text-to-SQL Challenge Towards Cross-Domain Natural Language Interfaces to Databases},
  author = {Tao Yu and Rui Zhang and Heyang Er and Suyi Li and Eric Xue and Bo Pang and Xi Victoria Lin and Yi Chern Tan and Tianze Shi and Zihan Li and Youxuan Jiang and Michihiro Yasunaga and Sungrok Shim and Tao Chen and Alexander Fabbri and Zifan Li and Luyao Chen and Yuwen Zhang and Shreya Dixit and Vincent Zhang and Caiming Xiong and Richard Socher and Walter S. Lasecki and Dragomir Radev},
  year = {2019},
  url = {https://aclanthology.org/D19-1204/},
  note = {10.18653/v1/D19-1204},
  booktitle = {EMNLP-IJCNLP},
  doi = {10.18653/v1/D19-1204},
}

@inproceedings{li2023bird,
  title = {Can LLM Already Serve as A Database Interface? A BIg Bench for Large-Scale Database Grounded Text-to-SQLs},
  author = {Jinyang Li and Binyuan Hui and Ge Qu and Jiaxi Yang and Binhua Li and Bowen Li and Bailin Wang and Bowen Qin and Rongyu Cao and Ruiying Geng and Nan Huo and Xuanhe Zhou and Chenhao Ma and Guoliang Li and Kevin C. C. Chang and Fei Huang and Reynold Cheng and Yongbin Li},
  year = {2023},
  url = {https://arxiv.org/abs/2305.03111},
  note = {arXiv:2305.03111},
  booktitle = {Advances in Neural Information Processing Systems 36 (Datasets and Benchmarks Track)},
  doi = {10.52202/075280-1835},
}

@article{wang2018executionguided,
  title = {Robust Text-to-SQL Generation with Execution-Guided Decoding},
  author = {Chenglong Wang and Kedar Tatwawadi and Marc Brockschmidt and Po-Sen Huang and Yi Mao and Oleksandr Polozov and Rishabh Singh},
  year = {2018},
  url = {https://arxiv.org/abs/1807.03100},
  note = {arXiv:1807.03100},
  journal = {arXiv},
}

@inproceedings{guo2019irnet,
  title = {Towards Complex Text-to-SQL in Cross-Domain Database with Intermediate Representation},
  author = {Jiaqi Guo and Zecheng Zhan and Yan Gao and Yan Xiao and Jian-Guang Lou and Ting Liu and Dongmei Zhang},
  year = {2019},
  url = {https://aclanthology.org/P19-1444/},
  booktitle = {Proceedings of the 57th Annual Meeting of the Association for Computational Linguistics},
  pages = {4524--4535},
  doi = {10.18653/v1/P19-1444},
}

@inproceedings{wang2020ratsql,
  title = {RAT-SQL: Relation-Aware Schema Encoding and Linking for Text-to-SQL Parsers},
  author = {Bailin Wang and Richard Shin and Xiaodong Liu and Oleksandr Polozov and Matthew Richardson},
  year = {2020},
  url = {https://aclanthology.org/2020.acl-main.677/},
  note = {10.18653/v1/2020.acl-main.677},
  booktitle = {ACL},
  doi = {10.18653/v1/2020.acl-main.677},
}

@inproceedings{scholak2021picard,
  title = {PICARD: Parsing Incrementally for Constrained Auto-Regressive Decoding from Language Models},
  author = {Torsten Scholak and Nathan Schucher and Dzmitry Bahdanau},
  year = {2021},
  url = {https://aclanthology.org/2021.emnlp-main.779/},
  note = {10.18653/v1/2021.emnlp-main.779},
  booktitle = {EMNLP},
  doi = {10.18653/v1/2021.emnlp-main.779},
}

@inproceedings{yao2023react,
  title = {ReAct: Synergizing Reasoning and Acting in Language Models},
  author = {Shunyu Yao and Jeffrey Zhao and Dian Yu and Nan Du and Izhak Shafran and Karthik Narasimhan and Yuan Cao},
  year = {2023},
  url = {https://arxiv.org/abs/2210.03629},
  note = {arXiv:2210.03629},
  booktitle = {ICLR},
}

@inproceedings{schick2023toolformer,
  title = {Toolformer: Language Models Can Teach Themselves to Use Tools},
  author = {Timo Schick and Jane Dwivedi-Yu and Roberto Dessi and Roberta Raileanu and Maria Lomeli and Luke Zettlemoyer and Nicola Cancedda and Thomas Scialom},
  year = {2023},
  url = {https://arxiv.org/abs/2302.04761},
  note = {arXiv:2302.04761},
  booktitle = {NeurIPS},
}

@article{gebru2021datasheets,
  title = {Datasheets for Datasets},
  author = {Timnit Gebru and Jamie Morgenstern and Briana Vecchione and Jennifer Wortman Vaughan and Hanna Wallach and Hal Daume III and Kate Crawford},
  year = {2021},
  url = {https://dl.acm.org/doi/10.1145/3458723},
  note = {10.1145/3458723},
  journal = {Communications of the ACM},
  doi = {10.1145/3458723},
}

@misc{openai2026structuredoutputs,
  title = {Structured model outputs},
  author = {OpenAI},
  year = {2026},
  url = {https://developers.openai.com/api/docs/guides/structured-outputs},
  note = {official documentation},
  howpublished = {OpenAI API documentation},
}

@misc{openai2026functioncalling,
  title = {Function calling},
  author = {OpenAI},
  year = {2026},
  url = {https://developers.openai.com/api/docs/guides/function-calling},
  note = {official documentation},
  howpublished = {OpenAI API documentation},
}
\end{document}